\pdfoutput=1

\documentclass[11pt]{article}

\usepackage{acl}

\usepackage{times}
\usepackage{latexsym}

\usepackage[T1]{fontenc}

\usepackage[utf8]{inputenc}

\usepackage{microtype}
\usepackage{natbib}

\usepackage{booktabs}
\usepackage{enumitem}
\usepackage{graphicx}
\usepackage{multirow}
\usepackage{tabularx}
\usepackage[normalem]{ulem}
\usepackage{CJKutf8}
\usepackage{float}
\usepackage{svg}
\usepackage{relsize}

\makeatletter
\def\uwave{\bgroup \markoverwith{\lower3.5\p@\hbox{\sixly \textcolor{red}{\char58}}}\ULon}
\makeatother

\title{As Little as Possible, as Much as Necessary: \\Detecting Over- and Undertranslations with Contrastive Conditioning}

\author{Jannis Vamvas$^1$ \and Rico Sennrich$^{1,2}$\\
  $^1$Department of Computational Linguistics, University of Zurich\\
  $^2$School of Informatics, University of Edinburgh \\ \medskip
  \texttt{\{vamvas,sennrich\}@cl.uzh.ch}}

\begin{document}
\maketitle
\begin{abstract}
Omission and addition of content is a typical issue in neural machine translation.
We propose a method for detecting such phenomena with off-the-shelf translation models.
Using contrastive conditioning, we compare the likelihood of a full sequence under a translation model to the likelihood of its parts, given the corresponding source or target sequence.
This allows to pinpoint superfluous words in the translation and untranslated words in the source even in the absence of a reference translation.
The accuracy of our method is comparable to a supervised method that requires a custom quality estimation model.

\end{abstract}

\section{Introduction}
Neural machine translation (NMT) is susceptible to coverage errors such as the addition of superfluous target words or the omission of important source content.
Previous approaches to detecting such errors make use of reference translations~\cite{yang2018otem}
or employ a separate quality estimation~(QE) model trained on synthetic data for a language pair~\cite{tuan-etal-2021-quality, zhou-etal-2021-detecting}.

In this paper, we propose a reference-free algorithm based on hypothetical reasoning.
Our premise is that a translation has optimal coverage if it uses \textit{as little information as possible and as much information as necessary} to convey the source sequence.
Therefore, an addition error means that the source would be better conveyed by a translation containing less information.
Conversely, an omission error means that the translation would be more adequate for a less informative source sequence.

\begin{figure*}
  \centering
  \includegraphics[width=\linewidth]{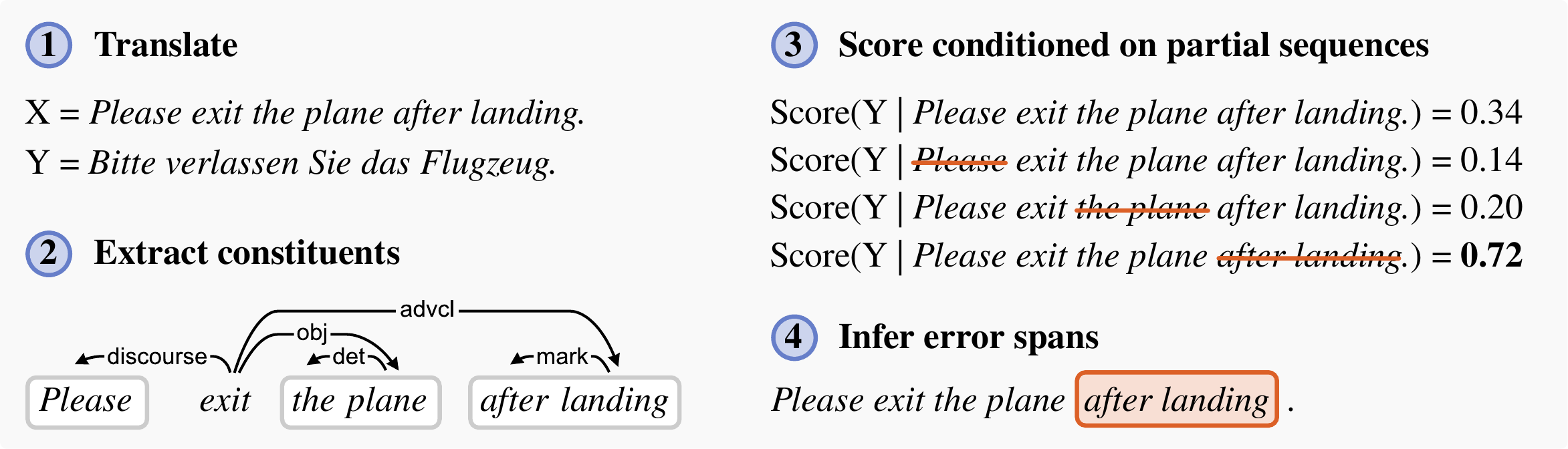}
  \caption{
  \label{fig:example}
  Example of how an omission error is detected.
  German translation~Y leaves \textit{after landing} erroneously untranslated (Step 1).
  Potential error spans are derived from a parse tree (Step~2).
  An NMT model such as mBART50 assigns a higher probability score to~Y conditioned on the source with \textit{after landing} deleted than to~Y conditioned on the full source (Step~3). This indicates that there is an omission error (Step~4).
  }
\end{figure*}

Adapting our \textit{contrastive conditioning} approach~\cite{vamvas-etal-2021-contrastive}, we use probability scores of NMT models to approximate this concept of coverage.
We create parse trees for both the source sequence and the translation, and treat their constituents as units of information.
Omission errors are detected by systematically deleting constituents from the source and by estimating the probability of the translation conditioned on such a partial source sequence.
If the probability score is higher than when the translation is conditioned on the full source, the deleted constituent might have no counterpart in the translation (Figure~\ref{fig:example}).
We apply the same principle to the detection of addition errors by swapping the source and the target sequence.

When comparing the detected errors to human annotations of coverage errors on the segment level~\cite{freitag2021experts}, our approach surpasses a supervised QE baseline that was trained on a large number of synthetic coverage errors.
Human raters find that word-level precision is higher for omissions than additions, with 39\% of predicted error spans being precise for English--German translations, and 20\% for Chinese--English.
False positive predictions can occur especially in cases where the translation has different syntax than the source.
We believe our algorithm could be a useful aid whenever humans remain in the loop, for example in a post-editing workflow.

We release the code and data to reproduce our findings, including a large-scale dataset of synthetic coverage errors in English–German and Chinese–English machine translations.\footnote{\url{https://github.com/ZurichNLP/coverage-contrastive-conditioning}}

\section{Related Work}

\paragraph{Coverage errors in NMT}
Addition and omission of target words have been observed by human evaluation studies in various languages, with omission as the more frequent error type~\cite{castilho2017a, 10.1162/tacl_a_00011}.
They are included as typical translation issues in the Multidimensional Quality Metrics (MQM) framework~\cite{lommel2014multidimensional}.
\textit{Addition} is defined as an accuracy issue where the target text includes text not present in the source, and \textit{omission} is defined as an accuracy issue where content is missing from the translation but is present in the source.\footnote{The terms \textit{overtranslation} and \textit{undertranslation} have been used in the literature as well. MQM reserves these terms for errors where the translation is too specific or too unspecific.}

\citet{freitag2021experts} used MQM to manually re-annotate English–German and Chinese–English machine translations submitted to the WMT 2020 news translation task~\cite{barrault-etal-2020-findings}.
Their findings confirm that state-of-the-art NMT systems still erroneously add and omit target words, and that omission occurs more often than addition.
Similar patterns can be found in English–French machine translations that have been annotated with fine-grained MQM labels for the document-level QE shared task~\cite{specia-etal-2018-findings, fonseca-etal-2019-findings, specia-etal-2020-findings-wmt}.

\paragraph{Detecting and reducing coverage errors}
While reference-based approaches include measuring the n-gram overlap to the reference~\cite{yang2018otem} and analyzing word alignment to the source~\cite{kong2019neural}, this work focuses on the \textit{reference-free} detection of coverage errors.

Previous work has employed custom QE models trained on labeled parallel data.
For example, \citet{zhou-etal-2021-detecting} insert synthetic hallucinations and train a Transformer to predict the inserted spans.
Similarly, \citet{tuan-etal-2021-quality} train a QE model on synthetically noisy translations.
In this paper, we propose a method that is based on off-the-shelf NMT models only.

Other related work has focused on improving coverage during decoding or training, for example via attention (\citealp{tu-etal-2016-modeling, wu2016google, li-etal-2018-simple}; among others).
More recently, \citet{yang-etal-2019-reducing} found that contrastive fine-tuning on references with synthetic omissions reduces coverage errors produced by an NMT system.

\section{Approach}

\paragraph{Contrastive Conditioning}
Properties of a translation can be inferred by estimating its probability conditioned on contrastive source sequences~\cite{vamvas-etal-2021-contrastive}.
For example, if a certain translation is more probable under an NMT model when conditioned on a counterfactual source sequence, the translation might be inadequate.

\paragraph{Application to Omission Errors}
Figure~\ref{fig:example} illustrates how contrastive conditioning can be directly applied to the detection of omission errors.
We construct \textit{partial source sequences} by systematically deleting constituents from the source.
If the probability score of the translation (average token log-probability) is higher when conditioned on such a partial source, the deleted constituent is taken to be missing from the translation.

To compute the probability score
for a translation~$Y$ given a source sequence~$X$,
we sum up the log-probabilities for every target token and normalize the sum by the number of target tokens:
\[
    \textrm{score}(Y|X)=\frac{1}{|Y|}\sum_{i=0}^{|Y|}\log p_{\theta}(y_{i}|X,y_{<i})
\]

\paragraph{Application to Addition Errors}
We apply the same method to addition detection, but swap the source and target languages.
Namely, we use an NMT model for the reverse translation direction, and we score the source sequence conditioned on the full translation and a set of partial translations.\footnote{Another possibility would be to leave the translation direction unreversed and to score the partial translations conditioned on the source. However, the scores might be confounded by a lack of fluency in the partial translations.
}

\paragraph{Potential Error Spans}
In its most basic form, our algorithm does not require any linguistic resources apart from tokenization.
For a source sentence of \textit{n} tokens one could create \textit{n} partial source sequences with the \textit{i}th token deleted.
However, such an approach would rely on a radical assumption of compositionality, treating all tokens as independent constituents.

We thus propose to extract potential error spans from parse trees, specifically from dependency trees predicted by Universal Dependency parsers~\cite{10.1162/coli_a_00402}, which are widely available.
This allows (a) to skip function words and (b) to include a reasonable number of multi-word spans in the set of potential error spans.
Formally, we consider word spans that satisfy the following conditions:
\begin{enumerate}[nosep]
  \item A potential error span is a complete subtree of the dependency tree.
  \item It covers a contiguous subsequence.
  \item It contains a part of speech of interest.
\end{enumerate}
For every potential error span, we create a partial sequence by deleting the span from the original sequence.
This is still a simplified notion of constituency, since some partial sequences will be ungrammatical.
Our assumption is that NMT models can produce reliable probability estimates despite the ungrammatical input.

\begin{table*}[]
\begin{tabularx}{\textwidth}{@{}lXrrrrrr@{}}
\toprule
 & Approach                 & \multicolumn{3}{l}{Detection of additions} & \multicolumn{3}{l}{Detection of omissions} \\
\textit{}                                 &                          & \multicolumn{1}{l}{Precision}       & \multicolumn{1}{l}{Recall}       & \multicolumn{1}{l}{F1}        & \multicolumn{1}{l}{Precision}       & \multicolumn{1}{l}{Recall}       & \multicolumn{1}{l}{F1}        \\ \midrule
\multirow{2}{*}{\textit{EN–DE}}  & Supervised baseline      & 6.9$\pm$1.9 & 2.9$\pm$0.9 & 4.0$\pm$1.3 & 40.3$\pm$5.2 & 6.1$\pm$0.1 & 10.6$\pm$0.2      \\
                                          & Our approach & 4.0\phantom{$\pm$0.0} & 15.0\phantom{$\pm$0.0} & \textbf{6.3}\phantom{$\pm$0.0} & 22.3\phantom{$\pm$0.0} & 18.8\phantom{$\pm$0.0} & \textbf{20.4}\phantom{$\pm$0.0}      \\ \midrule
\multirow{2}{*}{\textit{ZH–EN}} & Supervised baseline      & 4.3$\pm$0.6 & 4.7$\pm$0.7 & \textbf{4.5}$\pm$0.6 & 49.6$\pm$0.6 & 9.4$\pm$1.0 & 15.9$\pm$1.4      \\
                                          & Our approach & 1.7\phantom{$\pm$0.0} & 40.6\phantom{$\pm$0.0} & 3.4\phantom{$\pm$0.0} & 25.8\phantom{$\pm$0.0} & 62.0\phantom{$\pm$0.0} & \textbf{36.5}\phantom{$\pm$0.0} \\ \bottomrule
\end{tabularx}
\caption{Segment-level comparison of coverage error detection methods on the gold dataset by \citet{freitag2021experts}. We average over three baseline models trained with different random seeds, reporting the standard deviation.}
\label{tab:gold-standard-results}
\end{table*}

\section{Experimental Setup}\label{sec:experimental-setup}

In this section we describe the data and tools that we use to implement and evaluate our approach.

\paragraph{Scoring model}
We use mBART50~\cite{tang2020multilingual}, which is a sequence-to-sequence Transformer pre-trained on monolingual corpora in many languages using the BART objective~\cite{lewis-etal-2020-bart, 10.1162/tacl_a_00343} that was fine-tuned on English-centric multilingual MT in 50 languages.
Sequence-level probability scores are computed by averaging
the log-probabilities of all target tokens.
We use the one-to-many mBART50 model if English is the source language, and the many-to-one model if English is the target language.

\paragraph{Error spans}
We use Stanza~\cite{qi-etal-2020-stanza} for dependency parsing, a neural pipeline for various languages trained on data from Universal Dependencies~\cite{10.1162/coli_a_00402}.
We make use of universal part-of-speech tags (UPOS) to define parts of speech that might constitute potential error spans.
Specifically, we treat common nouns, proper nouns, main verbs, adjectives, numerals, adverbs, and interjections as relevant parts of speech.

\paragraph{Gold Standard Data}
We use state-of-the-art English–German and Chinese–English machine translations for evaluation, which have been annotated by~\citet{freitag2021experts} with translation errors.\footnote{\url{https://github.com/google/wmt-mqm-human-evaluation}}
We set aside translations by the system \textit{Online-B} as a development set, and use the other systems as a test set, excluding translations by humans.
The development set was used to identify the typical parts-of-speech of coverage error spans, listed in the paragraph above.

\paragraph{Synthetic Data}

\begin{figure}[]
  \centering
  \includesvg[width=\linewidth,pretex=\relscale{1.0}]{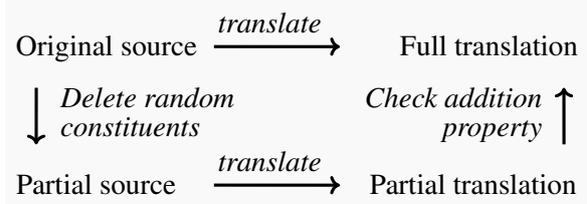}
  \caption{
    Process designed for creating machine translations with synthetic coverage errors.
    The full translation contains an addition error with regard to the partial source, and the partial translation contains an omission error with regard to the original source sequence.
  }
  \label{fig:synthetic-data-creation}
\end{figure}

We also create synthetic coverage errors, which we use for training a supervised baseline QE system.
We propose a data creation process that is inspired by previous work~\cite{yang-etal-2019-reducing,zhou-etal-2021-detecting,tuan-etal-2021-quality} but is defined such that it works for both additions and omissions, and produces fluent translations.

Figure~\ref{fig:synthetic-data-creation} illustrates the process.
We start from the original source sentences and create \textit{partial sources} by deleting randomly selected constituents.
Specifically, we delete each constituent with a probability of~15\%.
We then machine-translate both the original and the partial sources, yielding \textit{full} and \textit{partial machine translations}.
We retain only samples where the full machine translation is different from the partial one, and can be constructed by addition.

This allows us to treat the full translations as overtranslations of the partial sources, and the added words as addition errors.
Conversely, the partial translations are treated as undertranslations of the original sources.
Negative examples are created by pairing the original sources with the full translations, and the partial sources with the partial translations.\footnote{
Note that the synthetic dataset does not contain translations with both an addition and an omission error, which is a limitation.
Still, we expect that a system trained on the dataset will be able to generalize to such examples, especially if two separate classifiers are used for additions and omissions.
}

Our synthetic data are based on monolingual news text released for WMT.\footnote{\url{http://data.statmt.org/news-crawl/}}
To train the baseline system, we use 80k unique source segments per language pair. Statistics are reported in Table~\ref{tab:dataset-statistics-synthetic}.

\paragraph{Supervised baseline system}
Following the approach outlined by \citet{moura-etal-2020-ist}, we use the OpenKiwi framework~\cite{kepler-etal-2019-openkiwi} to train a separate Predictor-Estimator model~\cite{kim-etal-2017-predictor} per language pair, based on XLM-RoBERTa~\cite{conneau-etal-2020-unsupervised}.
The supervised task can be described as token-level binary classification.
Every token is classified as either \texttt{OK} or \texttt{BAD}, similar to the word-level labels used for the QE shared tasks~\cite{specia-etal-2020-findings-wmt}.
A source token is \texttt{BAD} if it is omitted in the translation, and a token in the translation is \texttt{BAD} if it is part of an addition error.
For English and German, we use the Moses tokenizer~\cite{koehn-etal-2007-moses} to separate the text into labeled tokens; for Chinese we label the text on the character level.

Where suitable, we use the default settings of OpenKiwi. We fine-tune the large version of XLM-RoBERTa, which results in a model of similar parameter count as the mBART50 model we use for contrastive conditioning.
We train for 10 epochs with a batch size of 32, with early stopping on the validation set.
For token classification we train two linear layers, separately for source and target language (which corresponds to omissions and additions, respectively).
We use AdamW~\cite{loshchilov2018decoupled} with a learning rate of 1e-5, freezing the pretrained encoder for the first 1000 steps.

\section{Evaluation}

\subsection{Segment-Level Comparison to Gold Data}\label{sec:mqm-evaluation}
The accuracy of our approach can be estimated based on the human ratings by \citet{freitag2021experts}.

\paragraph{Evaluation Design}
We use the MQM error types \textit{Accuracy/Addition} and \textit{Accuracy/Omission}, and ignore other types such as \textit{Accuracy/Mistranslation}.
We count a prediction as correct if any one of the human raters has marked the same error type anywhere in the segment.\footnote{We perform a segment-level evaluation and do not quantify word-level accuracy in this section since the dataset does not contain consistently annotated spans for coverage errors.}
We exclude segments from the evaluation that
might have been incompletely annotated (because raters stopped after marking five errors).
For ease of implementation, we also exclude segments that consist of multiple sentences.

\paragraph{Results}

The results of the gold-standard comparison are shown in Table~\ref{tab:gold-standard-results}.
Our approach clearly surpasses the baseline in the detection of omission errors in both language pairs.
However, both approaches recognize addition errors with low accuracy, and especially the supervised baseline has low recall. Considering its high performance on a synthetic test set~(Table~\ref{tab:synthetic-results} in the Appendix), it seems that the model does not generalize well to real-world coverage errors, highlighting the challenges of training a supervised QE model on purely synthetic data.

\subsection{Human Evaluation of Precision}\label{sec:human-evaluation}

\begin{table}[]
\begin{tabularx}{\columnwidth}{@{}lXrr@{}}
\toprule
& & EN–DE & ZH–EN \\ \midrule
\multirow{2}{*}{\textit{Target}}
& Addition errors   &  2.3  &  1.2   \\
 & Any errors       &  7.4  & 12.0  \\ \midrule
\multirow{2}{*}{\textit{Source}}
& Omission errors   & 36.3  & 13.8  \\
 & Any errors       & 39.4  & 19.5  \\ \bottomrule
\end{tabularx}
\caption{Human evaluation: word-level precision of the spans that were highlighted by our approach.
}
\label{tab:human-evaluation-main-results}
\end{table}

We perform an additional word-level human evaluation to analyze the predictions obtained via our approach in more detail.
Our human raters were presented segments that had been marked as true or false positives in the above evaluation, allowing us to quantify word-level precision.

\paragraph{Evaluation Design}
We employed two linguistic experts per language pair as raters.\footnote{Raters were paid ca.~USD~30 per hour.}
Each rater was shown around 700 randomly sampled positive predictions across both types of coverage errors.

Raters were shown the source sequence, the machine translation, and the predicted error span.
They were asked whether the highlighted span was indeed translated badly, and were asked to perform a fine-grained analysis based on a list of predefined answer options~(Figures \ref{fig:results-labels-additions} and \ref{fig:results-labels-omissions} in the Appendix).

A part of the samples were annotated by both raters.
The agreement was moderate for the main question, with a Cohen's kappa of 0.54 for English–German and 0.45 for Chinese–English.
Agreement on the more subjective follow-up question was lower (0.32 / 0.13).

\paragraph{Results}
The fine-grained answers allow us to quantify the word-level precision of the spans highlighted by our approach, both with respect to coverage errors in particular and to translation errors in general~(Table~\ref{tab:human-evaluation-main-results}).
Precision is higher than expected when detecting omission errors in English–German translations, but is still low for additions.
The distribution of the detailed answers~(Figures~\ref{fig:results-labels-additions} and~\ref{fig:results-labels-omissions} in the Appendix) suggests that syntactical differences between the source and target language contribute to the false positives regarding additions.
Example predictions are provided in Appendix~\ref{sec:examples}, which include cases where all three raters of~\citet{freitag2021experts} had overlooked the coverage error.

Finally, Table~\ref{tab:human-evaluation-main-results} shows that many of the predicted error spans are in fact translation errors, but not coverage errors in a narrow sense.
For example, more than 10\% of the spans marked in Chinese--English translations were classified by our raters as a different type of accuracy error, such as mistranslation.

\section{Limitations and Future Work}
We hope that the automatic detection of coverage errors could be an aid to translators and post-editors, given that manually detecting such errors is tedious.
Our results on omissions are encouraging, and user studies are recommended in order to validate the usefulness of the predictions to practitioners.
Further work needs to be done to improve the detection of additions, of which the real-world data contain few examples.
Higher accuracy would be necessary for word-level QE to be helpful~\cite{shenoy-etal-2021-investigating}, and so with regard to detecting addition errors, the practical utility of both the baseline and of our approach remains limited.

Inference time should also be discussed.
In Appendix~\ref{sec:inference-time} we perform a comparison, finding that on a long sentence pair contrastive conditioning can take up to ten times longer than a forward pass of the baseline.
However, this is still a fraction of the time needed for generating a translation in the first place.
In addition, restricting the potential error spans that are considered could further improve efficiency.

\section{Conclusion}
We have proposed a reference-free method to automatically detect coverage errors in translations.
Derived from contrastive conditioning, our method relies on hypothetical reasoning over the likelihood of partial sequences.
Since any off-the-shelf NMT model can be used to estimate conditional likelihood, no access to the original translation system or to a quality estimation model is needed.
Evaluation on real machine translations shows that our approach outperforms a supervised baseline in the detection of omissions.
Future work could address the low precision on addition errors, which are relatively rare in the datasets we used for evaluation.

\section*{Acknowledgments}
This work was funded by the Swiss National Science Foundation (project MUTAMUR; no.~176727).
We would like to thank Xin Sennrich for facilitating the recruitment of annotators, and Chantal Amrhein as well as the anonymous reviewers for helpful feedback.


\bibliography{bibliography}
\bibliographystyle{acl_natbib}

\appendix

\setcounter{table}{0}
\renewcommand{\thetable}{A\arabic{table}}

\vfill

\section{Annotator Guidelines}\label{sec:annotator-guidelines}

\textit{You will be shown a series of source sentences and translations. One or several spans in the text are highlighted and it is claimed that the spans are translated badly. You are asked to determine whether the claim is true. The highlighted spans can be either in the source sequence or in the translation. If a span is in the source sentence, check whether it has been correctly translated. If a span is in the translation, check whether it correctly conveys the source.
Sometimes, multiple spans are highlighted. In that case, focus your answer on the span that is most problematic for the translation. In a second step, you are asked to select an explanation. On the one hand, if you agree that the highlighted span is translated badly, please explain your reasoning by selecting your explanation. On the other hand, if you disagree and think that the span is well-translated, please select an explanation why the span might have been marked as badly translated in the first place.
Should multiple explanations be equally plausible, select the first from the top.}

\begin{table*}[]
\begin{tabularx}{\textwidth}{@{}Xllllllll@{}}
\toprule
                                      & \multicolumn{4}{l}{Detection of additions} & \multicolumn{4}{l}{Detection of omissions} \\
                       & \textit{Prec.}       & \textit{Recall}       & \textit{F1} & \textit{MCC}       & \textit{Prec.}       & \textit{Recall}       & \textit{F1}     & \textit{MCC}      \\ \midrule
\textit{EN–DE} & \\
Supervised \\ \mbox{\phantom{––}Baseline} & 98.8$\pm$0.4 &           98.0$\pm$.2 &            \textbf{98.4}$\pm$.2 &            \textbf{96.8}$\pm$.1 &            94.0$\pm$1.3 &            96.6$\pm$0.4 &            \textbf{95.3}$\pm$.5 &            \textbf{90.5}$\pm$.2  \\
 Ours &     78.1\phantom{$\pm$0.0} & 88.3\phantom{$\pm$.0} &  82.9\phantom{$\pm$.0} &  76.7\phantom{$\pm$.0} &  80.9\phantom{$\pm$0.0} &  98.6\phantom{$\pm$0.0} &  88.9\phantom{$\pm$.0} &  78.1\phantom{$\pm$.0} \\ \midrule
\textit{ZH–EN} & \\
Supervised \\ \mbox{\phantom{––}Baseline}  &  87.2$\pm$1.5 &           75.7$\pm$.6 &            \textbf{81.0}$\pm$.3 &            \textbf{72.6}$\pm$.6 &            67.3$\pm$1.3 &            68.0$\pm$1.2 &            \textbf{67.7}$\pm$.9 &            \textbf{53.8}$\pm$.3     \\
 Ours &     26.1\phantom{$\pm$0.0} & 88.9\phantom{$\pm$.0} &  40.4\phantom{$\pm$.0} &  23.3\phantom{$\pm$.0} &  28.3\phantom{$\pm$0.0} &  92.0\phantom{$\pm$0.0} &  43.3\phantom{$\pm$.0} &  40.3\phantom{$\pm$.0} \\ \bottomrule
\end{tabularx}
\caption{Segment-level and word-level (\textit{MCC}) evaluation based on a test set with synthetic coverage errors.}
\label{tab:synthetic-results}
\end{table*}

\begin{table*}[]
\vspace{0.5cm}
\begin{tabularx}{\textwidth}{@{}Xr@{\hskip 0.3in}r@{\hskip 0.3in}r@{\hskip 0.5in}r@{\hskip 0.3in}r@{\hskip 0.3in}r@{}}
\toprule
                   & \multicolumn{3}{l}{Short sentence pair} & \multicolumn{3}{@{}l}{Long sentence pair} \\ \midrule
                   & Additions     & Omissions     & Both    & Additions     & Omissions    & Both    \\
\mbox{Supervised baseline}           & -             & -             & 25\,ms      & -             & -            & 25\,ms      \\
Our approach       & 40\,ms            & 45\,ms            & 83\,ms      & 165\,ms           & 197\,ms          & 365\,ms     \\
– excluding parser & 18\,ms            & 21\,ms            & 38\,ms      & 102\,ms           & 144\,ms          & 239\,ms     \\ \bottomrule
\end{tabularx}
\caption{Inference times when predicting on a short and a long sentence pair. Since we did not use a parser that is optimized for efficiency, we additionally report inference time without including the time needed for parsing.}
\label{tab:inference-time}
\vspace{0.3cm}
\end{table*}

\pagebreak

\section{Evaluation on Synthetic Errors}\label{sec:evaluation-on-synthetic-errors}

We used a test split held back from the synthetic data to perform an additional evaluation.
On the segment level, we report Precision, Recall and F1-score.
Like in Section \ref{sec:mqm-evaluation}, a prediction is treated as correct on the segment level if for a predicted coverage error there is indeed a coverage error of that type anywhere in the segment.

On the word level, we follow previous work on word-level QE \cite{specia-etal-2020-findings-wmt} and report the Matthews correlation coefficient (MCC) across all the tokens in the test set.

\paragraph{Results}
Results are shown in Table~\ref{tab:synthetic-results}.
The supervised baseline has a high accuracy on English–German translations and a moderate accuracy on Chinese–English translations.
In comparison, our approach performs clearly worse than the supervised baseline on the synthetic errors.

\vfill

\section{Inference Time}\label{sec:inference-time}
Inference times are reported in Table~\ref{tab:inference-time}.
We measure the time needed to run the coverage error detection methods on a short sentence pair and on a long sentence pair for English–German.
The short sentence pair is taken from Figure~\ref{fig:example} and the long sentence pair has 40~tokens in the source sequence and 47~tokens in the target sequence.
We average over~1000 repetitions on RTX 2080 Ti GPUs.

The higher inference times for our approach can be explained by the number of translation probabilities that need to be estimated.
On average, we compute 30~scores per sentence in the English–German MQM dataset, and 44~per sentence in the Chinese–English MQM dataset.
Still, the time needed for computing all these scores is only a fraction of the time it takes to generate a translation (254\,ms for the short source sentence and~861\,ms for the long sentence, assuming a beam size of~5).

The required number of scores could be reduced by considering fewer potential error spans.
Furthermore, scoring could be parallelized across batches of multiple translations.
Finally, using a more efficient parser, or no parser at all, could speed up inference.

\vfill
\pagebreak
\onecolumn

\section{Dataset Statistics}
\begin{table*}[htb!]
\begin{tabularx}{\textwidth}{@{}Xrrrrrrr@{}}
\toprule
Dataset split & \multicolumn{3}{l}{Number of segments}                        & \multicolumn{4}{l}{Number of tokens}             \\
              & \multicolumn{1}{l}{Total}              & \multicolumn{1}{l}{W/ addition} & \multicolumn{1}{l}{W/ omission} & \multicolumn{1}{l}{Src. \texttt{OK}}        & \multicolumn{1}{l}{Src. \texttt{BAD}} & \multicolumn{1}{l}{Tgt. \texttt{OK}} & \multicolumn{1}{l}{Tgt. \texttt{BAD}} \\ \midrule
EN–DE Train   & 135269                                 & 18423                                      & 18423                                      & 2185918                              & 58378                          & 2197843                       & 53911                          \\
EN–DE Dev     & 16984                                  & 2328                                       & 2328                                       & 273311                               & 7398                           & 275156                        & 6781                           \\
EN–DE Test    & 16984                                  & 2328                                       & 2328                                       & 273277                               & 7701                           & 275036                        & 7032                           \\ \midrule
ZH–EN Train   & 110195                                 & 10697                                      & 10697                                      & 2576135                              & 62311                          & 1866567                       & 37730                          \\
ZH–EN Dev     & 14149                                  & 1383                                       & 1383                                       & 326743                               & 7562                           & 236685                        & 4244                           \\
ZH–EN Test    & 14026                                  & 1342                                       & 1342                                       & 322000                               & 7566                           & 234757                        & 4882                           \\ \bottomrule
\end{tabularx}
\caption{Statistics for the dataset of synthetic coverage errors described in Section~\ref{sec:experimental-setup}.}
\label{tab:dataset-statistics-synthetic}
\end{table*}

\begin{table*}[htb!]
\begin{tabularx}{\textwidth}{@{}Xrrr@{}}
\toprule
  Dataset split                          & \multicolumn{3}{l}{Number of segments}                                                                              \\
                            & \multicolumn{1}{l}{Total} & \multicolumn{1}{l}{With an addition error} & \multicolumn{1}{l}{With an omission error} \\ \midrule
EN–DE Dev                   & 1418                      & 77                                         & 187                                        \\
EN–DE Test                  & 8508                      & 407                                        & 1057                                       \\
– without excluded segments & 4839                      & 162                                        & 484                                        \\ \midrule
ZH–EN Dev                   & 1999                      & 69                                         & 516                                        \\
ZH–EN Test                  & 13995                     & 329                                        & 3360                                       \\
– without excluded segments & 8851                      & 149                                        & 1569                                       \\ \bottomrule
\end{tabularx}
\caption{Statistics for the gold dataset by \citet{freitag2021experts}.}
\label{tab:dataset-statistics-mqm}
\end{table*}

\bigskip

\section{Examples of Synthetic Coverage Errors}\label{sec:examples-of-synthetic-coverage-errors}

\medskip
\noindent
\subsection*{English–German Example}
\textbf{Addition error}\\
\textit{Partial source:} But they haven't played.\\
\textit{Full machine translation:} Aber sie haben nicht \uwave{gegen ein Team wie uns} gespielt.

\smallskip
\noindent \textbf{Omission error}\\
\textit{Full source:} But they haven't played \uwave{against a team like us}.\\
\textit{Partial machine translation:} Aber sie haben nicht gespielt.

\bigskip
\noindent
\subsection*{Chinese–English Example}
\textbf{Addition error}\\
\textit{Partial source:} \begin{CJK*}{UTF8}{gbsn}医院和企业共同研发相关检测试剂盒，惠及更多患者。\end{CJK*}\\
\textit{Full translation:} Hospitals and enterprises jointly develop related test kits to benefit more \uwave{cancer} patients.

\smallskip
\noindent \textbf{Omission error}\\
\textit{Full source:} \begin{CJK*}{UTF8}{gbsn}医院和企业共同研发相关检测试剂盒，惠及更多\uwave{肿瘤}患者。\end{CJK*}\\
\textit{Partial translation:} Hospitals and enterprises jointly develop related test kits to benefit more patients.

\vfill
\pagebreak

\section{Examples of Coverage Errors Predicted by Contrastive Conditioning}\label{sec:examples}

\subsection*{English–German Examples}

\medskip
\noindent
\textbf{Predicted addition error}\\
\textit{Source:} He added: "It's backfired on him now, though, that's the sad thing."\\
\textit{Machine translation:} Er fügte \uwave{\textbf{hinzu}}: "Es ist jetzt auf ihn abgefeuert, aber das ist das Traurige."

\smallskip
\noindent
\textit{Original MQM rating~\cite{freitag2021experts}:} \textit{No related accuracy error marked by the three raters.}

\smallskip
\noindent
\textit{Answer by our human rater:} \textit{The highlighted target span is not translated badly. It might have been highlighted because it is syntactically different from the source.}

\smallskip
\noindent
\textit{Meaning of highlighted span:} hinzu = `additionally'

\medskip
\noindent
\textbf{Predicted omission error}\\
\textit{Source:} UK's medical \uwave{\textbf{drug}} supply still uncertain in no-deal Brexit\\
\textit{Machine translation:} Die medizinische Versorgung Großbritanniens ist im No-Deal-Brexit noch ungewiss

\smallskip
\noindent
\textit{Original MQM rating:} \textit{No accuracy error marked by the three raters.}

\smallskip
\noindent
\textit{Answer by our human rater:} \textit{The highlighted source span is indeed translated badly. It contains information that is
missing in the translation but can be inferred or is trivial.}

\bigskip
\noindent
\textbf{Predicted omission error}\\
\textit{Source:} The automaker is expected to report its quarterly vehicle deliveries in the next \uwave{\textbf{few}} days.\\
\textit{Machine translation:} Der Autohersteller wird voraussichtlich in den nächsten Tagen seine vierteljährlichen Fahrzeugauslieferungen melden.

\smallskip
\noindent
\textit{Original MQM rating:} \textit{No related accuracy error marked by the three raters.}

\smallskip
\noindent
\textit{Answer by our human rater:} \textit{The highlighted source span is not translated badly. The words in the span do not need to be translated.}

\medskip
\noindent
\subsection*{Chinese–English Examples}

\medskip
\noindent
\textbf{Predicted addition error}\\
\textit{Source:} \begin{CJK*}{UTF8}{gbsn}美方指责伊朗制造了该袭击，并对伊朗实施新制裁。\end{CJK*}\\
\textit{Machine translation:} The US accused Iran of causing the attack and imposed new sanctions \uwave{\textbf{on Iran}}.

\smallskip
\noindent
\textit{Original MQM rating~\cite{freitag2021experts}:} \textit{No related accuracy error marked by the three raters.}

\smallskip
\noindent
\textit{Answer by our human rater:} \textit{The highlighted target span is not translated badly. No phenomenon that might have caused the prediction was identified.}

\medskip
\noindent
\textbf{Predicted omission error}\\
\textit{Source:} \begin{CJK*}{UTF8}{gbsn}\uwave{目前}已收到来自俄罗斯农业企业的约50项申请。\end{CJK*}\\
\textit{Machine translation:} About 50 applications have been received from Russian agricultural enterprises.

\smallskip
\noindent
\textit{Original MQM rating:} \textit{No accuracy error marked by the three raters.}

\smallskip
\noindent
\textit{Answer by our human rater:} \textit{The highlighted source span is indeed translated badly. It contains information that is missing in the translation.}

\smallskip
\noindent
\textit{Meaning of highlighted span:} \begin{CJK*}{UTF8}{gbsn}目前\end{CJK*} = `at present'

\bigskip
\noindent
\textbf{Predicted omission error}\\
\textit{Source:} \begin{CJK*}{UTF8}{gbsn}他说，该系统目前在世界上有很大需求，但俄罗斯军队也需要它，\\\uwave{其中}包括在北极地区。\end{CJK*}\\
\textit{Machine translation:} He said that the system is currently in great demand in the world, but the Russian army also needs it, including in the Arctic.

\smallskip
\noindent
\textit{Original MQM rating:} \textit{No accuracy error marked by the three raters.}

\smallskip
\noindent
\textit{Answer by our human rater:} \textit{The highlighted source span is not translated badly. The words in the span do not need to be translated.}

\smallskip
\noindent
\textit{Meaning of highlighted span:} \begin{CJK*}{UTF8}{gbsn}其中\end{CJK*} = `among'

\vfill
\pagebreak

\section{Detailed Results of Human Evaluation}\label{sec:detailed-results-of-human-evaluation}

\begin{figure*}[htb!]
  \centering
  \includegraphics[width=\linewidth]{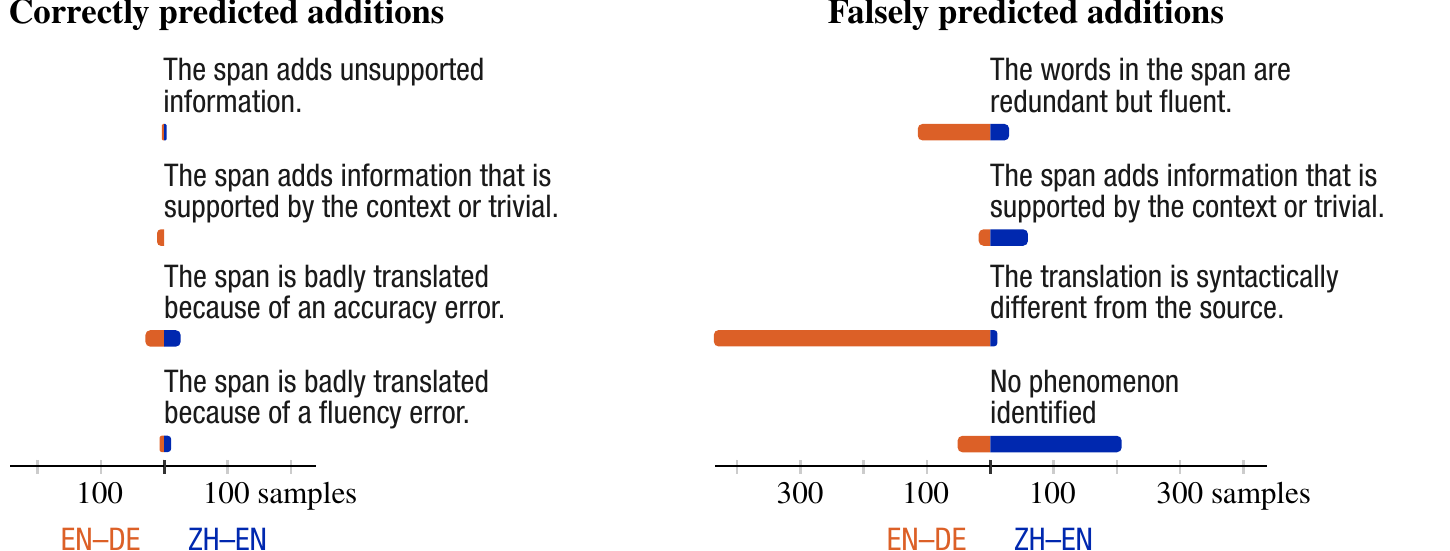}
  \caption{Results for the human evaluation of predicted addition errors. If human raters answered that the highlighted span in the translation was indeed badly translated, they were offered the four explanation options on the left. Otherwise they chose from the four options on the right.}
  \label{fig:results-labels-additions}
\end{figure*}

\begin{figure*}[htb!]
  \centering
  \includegraphics[width=\linewidth]{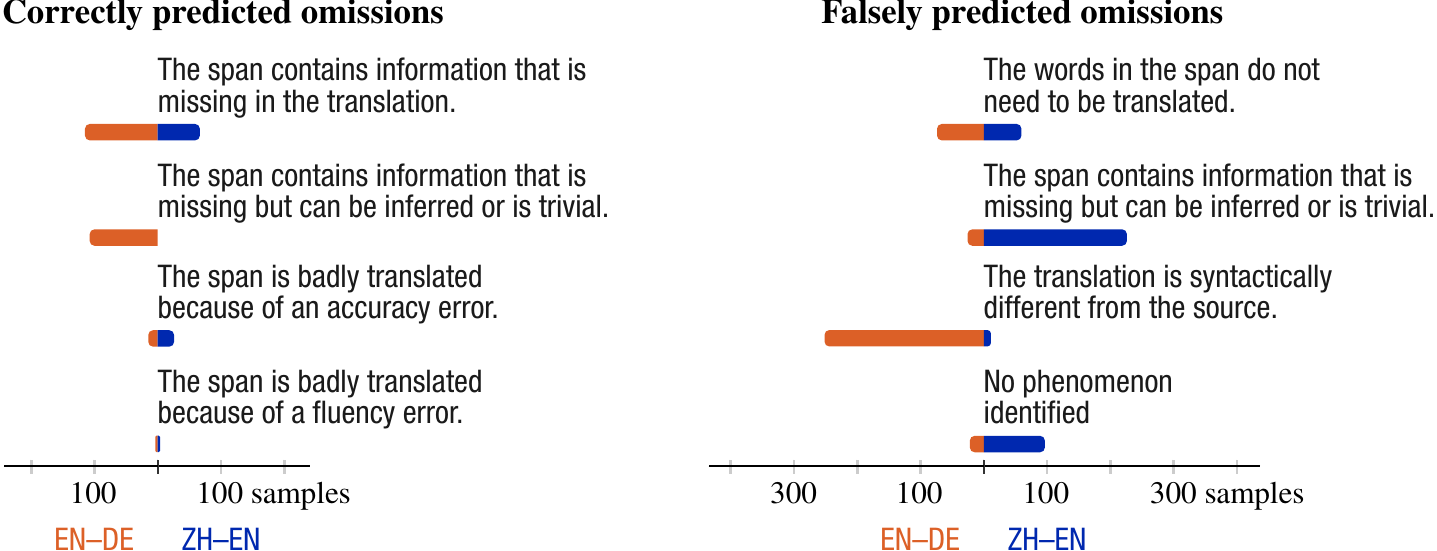}
  \caption{Results for the human evaluation of predicted omission errors. If human raters answered that the highlighted span in the source sequence was indeed badly translated, they were offered the four explanation options on the left. Otherwise they chose from the four options on the right.}
  \label{fig:results-labels-omissions}
\end{figure*}

\end{document}